\documentclass[conference]{IEEEtran}
\IEEEoverridecommandlockouts

\usepackage{cite}
\usepackage{amsmath,amssymb,amsfonts}
\usepackage{graphicx}
\usepackage{subcaption}
\usepackage{booktabs}
\usepackage{multirow}
\usepackage{textcomp}
\usepackage{xcolor}
\usepackage{float}
\usepackage{url}
\usepackage{microtype}
\usepackage{mathtools}
\usepackage{amsthm}
\usepackage[capitalize,noabbrev]{cleveref}

\theoremstyle{plain}

\theoremstyle{definition}

\theoremstyle{remark}

\def\BibTeX{{\rm B\kern-.05em{\sc i\kern-.025em b}\kern-.08em
    T\kern-.1667em\lower.7ex\hbox{E}\kern-.125emX}}

\begin{document}

\title{A Diagnostic Gap Framework for Evaluating\\
Reconstruction Fidelity in Weakly Supervised Mammography}

 \author{
 \IEEEauthorblockN{Vinceline Bertrand}
\IEEEauthorblockA{
 \textit{Florida Atlantic University} \\
 \textit{Boca Raton, FL, USA}
}
 \and
 \IEEEauthorblockN{Ionut Cardei}
 \IEEEauthorblockA{
 \textit{Florida Atlantic University} \\
 \textit{Boca Raton, FL, USA}
 }
 }

\maketitle

\begin{abstract}
Weakly supervised pipelines for medical imaging have become increasingly popular over the years. These systems often include multiple stages and components, such as reconstruction, generation, and localization, yet standard evaluation metrics provide limited insight into whether clinically relevant information is preserved across each stage. We present the diagnostic gap framework, a practical evaluation tool that measures decision preservation and explanation preservation as a function of measured reconstruction fidelity. To isolate the effect of reconstruction from localization, we evaluate on curated lesion ROI crops using a fidelity ladder of three class-conditional reconstructors---VQ-VAE-GAN, VAE-GAN, and diffusion (SDEdit)---spanning a twenty-fold range in perceptual distance (LPIPS 0.029--0.584). At autoencoder fidelity, both decision and explanation are preserved: AUC changes remain within $\pm$0.005 and attribution similarity (HiResCAM, Grad-CAM++) stays high. At diffusion fidelity, both collapse: pooled AUC drops by 0.253 and mass-pathology AUC falls below chance. The diagnostic gap is thus a measurable function of reconstruction fidelity rather than an intrinsic cost of reconstruction, and the framework provides an architecture-agnostic instrument for identifying when and where multi-stage pipelines lose diagnostic signal.
\end{abstract}


\section{Introduction}

Weakly supervised AI pipelines for medical imaging can be complex, spanning classification, attribution-based localization, and generative reconstruction into multi-stage systems~\cite{zhou2016cam,selvaraju2017gradcam}. Generative reconstruction is a common downstream component in such pipelines: after a classifier localizes a region of interest via attribution maps, a reconstructor produces a normalized representation of the lesion for standardized analysis, data augmentation, or visual explanation~\cite{oord2017vqvae,kazerouni2023diffusion_survey}. Yet the tools used to evaluate these pipelines---aggregate metrics such as accuracy, recall, or AUC---were designed for single-stage models and provide limited insight into where diagnostic information is lost~\cite{zech2018confounding}. This creates a real problem; when a pipeline's output degrades clinically, it becomes hard for us to tell whether the failure lies in the classifier, the localizer, or the reconstructor.

A central open question is whether observed diagnostic degradation is an inherent cost of passing images through a generative model, or whether it depends on the fidelity the reconstructor achieves. If degradation tracks fidelity, then the ``diagnostic gap'' is a controllable quantity (high-fidelity reconstructors incur little of it) rather than a fundamental limitation of reconstruction-based pipelines. Answering this question requires comparing reconstructors that span a wide fidelity range under otherwise identical conditions.

We address this with the diagnostic gap framework, a stage-aware evaluation tool built on two principles. First, the diagnostic classifier is frozen and serves as a fixed measurement instrument: any change in its output is attributable to reconstruction, not retraining. Second, we measure both decision preservation (whether the classifier's prediction changes) and explanation preservation (whether its attribution maps shift), because a reconstruction can leave the decision intact while altering the evidence the classifier relies on.

We apply this framework to mammography, where breast cancer screening demands reliable fine-grained reasoning despite subtle lesion morphology and tissue variability~\cite{american_cancer_society,chotai2026why}. To isolate the reconstruction question from localization artifacts, all analysis is performed on curated lesion regions of interest (ROI) crops. We compare three class-conditional reconstructors: VQ-VAE-GAN~\cite{oord2017vqvae,esser2021taming}, VAE-GAN~\cite{larsen2016autoencoding}, and diffusion via SDEdit~\cite{meng2021sdedit}. This forms a fidelity ladder that spans a twenty-fold range in perceptual distance.

\subsection{Contributions}
\begin{itemize}
  \item We introduce the diagnostic gap framework, combining a variety of metrics to evaluate reconstruction fidelity in a controlled, architecture-agnostic manner. At autoencoder fidelity (LPIPS $\leq$ 0.078), both decision and explanation are preserved: AUC changes remain within $\pm$0.005 and prediction agreement exceeds 96\% for lesion type. At diffusion fidelity (LPIPS 0.584), both collapse: pooled pathology AUC drops by 0.253 and mass-pathology AUC falls to 0.461.
  \item We construct a reconstruction fidelity ladder comparing discrete-latent (VQ-VAE-GAN), continuous-latent (VAE-GAN), and diffusion-based (SDEdit) reconstructors, spanning a twenty-fold range in perceptual distance (LPIPS 0.029--0.584). An SDEdit strength sweep across seven levels (0.05--0.7) confirms monotonic degradation: $\Delta$AUC decreases steadily from near zero to $-$0.292 as fidelity worsens, establishing that diagnostic degradation is fidelity-dependent rather than intrinsic to reconstruction.
  \item We evaluate explanation preservation using two modern attribution methods (HiResCAM and Grad-CAM++), showing that attribution fidelity is the more sensitive diagnostic indicator. At autoencoder fidelity, attribution similarity (HiResCAM cosine 0.78--0.84; Grad-CAM++ cosine 0.87--0.90) discriminates VAE-GAN from VQ-VAE-GAN where decision metrics saturate, while at diffusion fidelity both attribution and decision metrics collapse together.
\end{itemize}

\section{Related Work}
\label{sec:related}

Deep learning has achieved strong performance for breast cancer detection in mammography using CNNs and ensemble models~\cite{shen2019deep,yi2017breast}. However, evaluation still relies predominantly on aggregate classification metrics, which can conceal stage-specific losses in diagnostic information.

\textbf{Generative reconstruction in medical imaging.} Discrete-latent models such as Vector Quantized Variational Auto Encoders (VQ-VAEs)~\cite{oord2017vqvae} support efficient representation learning and are often combined with adversarial objectives~\cite{larsen2016autoencoding,esser2021taming}. Continuous-latent VAEs provide a natural comparison by removing the quantization bottleneck while retaining the same architectural capacity. Diffusion models~\cite{kazerouni2023diffusion_survey} offer superior mode coverage but typically operate at lower reconstruction fidelity when used for editing via SDEdit~\cite{meng2021sdedit}. Existing medical-image reconstruction work typically evaluates image quality using pixel-level and perceptual similarity metrics, such as PSNR, SSIM, and LPIPS \cite{li2025patientspecific}; we instead assess reconstruction through downstream diagnostic behavior.

\textbf{Attribution methods.} Gradient-based attribution methods including Grad-CAM~\cite{selvaraju2017gradcam} enable coarse localization but produce highly interpolated heatmaps that are upsampled to the input image size. Higher-resolution alternatives such as HiResCAM~\cite{draelos2021hirescam} and Grad-CAM++~\cite{chattopadhay2018gradcampp} provide tighter attribution maps. We use both to verify that our conclusions about explanation preservation are robust to the choice of attribution method.

\textbf{Calibration and uncertainty quantification.} Post-hoc calibration methods including temperature scaling~\cite{guo2017calibration} and Dirichlet calibration~\cite{kull2019dirichlet} improve the reliability of predicted probabilities without retraining. Conformal prediction~\cite{angelopoulos2021conformal} provides distribution-free coverage guarantees through prediction sets, enabling model-free uncertainty characterization. We use both to validate the diagnostic probe as a reliable measurement instrument before applying it to assess reconstruction quality.

\textbf{Evaluation tools for medical AI.} While extensive work has addressed model development, fewer tools exist for diagnosing \emph{where} multi-stage AI pipelines lose clinically relevant information. Recent stage-aware medical imaging architectures have shown that assigning different modeling strategies to different network stages can improve the balance between local detail preservation and semantic context~\cite{wang2026cmtunet}. However, these approaches evaluate the final model rather than measuring how clinically relevant information changes across a pipeline. Stage-wise evaluation has been explored for confounding detection~\cite{zech2018confounding}, and sanity checks for attribution methods~\cite{adebayo2018sanity} have revealed cases where saliency maps are independent of model parameters. However, systematic frameworks for measuring task-dependent diagnostic degradation as a function of reconstruction fidelity remain lacking. Our framework fills this gap by combining frozen-probe evaluation, dual decision--explanation measurement, and fidelity-indexed comparison into a reusable evaluation protocol.

\section{Methodology}
\label{sec:method}

We study how reconstruction fidelity affects the preservation of diagnostic information in mammography. Given a lesion ROI and a fixed classifier, we ask: does reconstruction change the classifier's \emph{decision}, and does it change the classifier's \emph{evidence}? We answer both as a function of measured reconstruction fidelity, using a ladder of reconstructors that spans a wide range of fideliy levels.

\subsection{Diagnostic Probe}

Diagnosis follows a two-stage cascade mirroring clinical reading. A lesion-type stage distinguishes calcification from mass; a lesion-type-specific pathology stage distinguishes benign from malignant, yielding three classifiers (probes): lesion-type, calcification-pathology, and mass-pathology. Each is a ResNet-50~\cite{he2016resnet} pretrained on ImageNet and fine-tuned on $256\times256$ ROI crops with class-weighted cross-entropy. Predictive reliability is assessed using temperature scaling~\cite{guo2017calibration}, Dirichlet calibration~\cite{kull2019dirichlet}, and score-based conformal prediction~\cite{angelopoulos2021conformal}. Temperature scaling is fit on the validation set after training, and the calibrated probes are then frozen for every reconstruction experiment. The same probe therefore evaluates all rungs of the fidelity ladder, ensuring a controlled comparison.

\subsection{Reconstruction Fidelity Ladder}

We compare three class-conditional reconstructors, each trained on the same ROI crops and conditioned on the lesion's label, chosen to occupy different points on the fidelity axis:

\textbf{VQ-VAE-GAN} (discrete latent): A conditional encoder-decoder with a single shared vector-quantized codebook (size 512, embedding dimension 256) updated by exponential moving average. A U-Net skip connection preserves spatial detail and the decoder is conditioned on a class embedding.

\textbf{VAE-GAN} (continuous latent): The identical architecture with the discrete codebook replaced by a Gaussian latent regularized by a KL term. This isolates the effect of the discrete bottleneck while holding capacity, conditioning, and the adversarial objective fixed.

\textbf{Diffusion (SDEdit)}: A class-conditional DDPM/DDIM U-Net~\cite{meng2021sdedit} with block channels (128, 128, 256, 256, 512, 512), two layers per block, and a single attention stage, implemented via Hugging Face \texttt{diffusers} \texttt{UNet2DModel} with class-conditional embedding. The model is trained with 1,000 diffusion timesteps. Reconstruction uses SDEdit: the ROI is partially noised to a specified strength and denoised with DDIM (50 steps). Larger strength regenerates more of the image and is less faithful; we sweep strength across seven levels (0.05, 0.1, 0.15, 0.2, 0.35, 0.5, 0.7) to trace preservation across a continuous fidelity range.

The VQ-VAE-GAN and VAE-GAN are trained with a two-phase schedule: reconstruction plus VGG perceptual loss, followed by PatchGAN adversarial refinement. The diffusion model is trained as a standard denoiser with EMA weights used at inference.

\subsection{Reconstruction Fidelity Metrics}

Fidelity between an original ROI and its reconstruction is quantified with LPIPS~\cite{zhang2018lpips} (VGG backbone; perceptual distance, lower is better) and SSIM (structural similarity, higher is better). These define the continuous fidelity axis used throughout the analysis.

\subsection{Decision Preservation}

On held-out test ROIs, the frozen probe classifies both the original crop and each reconstruction. We report four complementary measures:
\begin{itemize}
  \item \textbf{Prediction agreement}: fraction of ROIs whose predicted class is unchanged.
  \item \textbf{Accuracy change} ($\Delta\text{Acc}$): $\text{acc}_{\text{recon}} - \text{acc}_{\text{orig}}$.
  \item \textbf{AUC change} ($\Delta\text{AUC}$): $\text{AUC}_{\text{recon}} - \text{AUC}_{\text{orig}}$.
  \item \textbf{Probability shift} ($\overline{|\Delta p|}$): mean absolute change in predicted malignancy probability, capturing sub-threshold degradation that prediction agreement misses.
\end{itemize}
Values near zero indicate diagnostic preservation; negative $\Delta\text{AUC}$ quantifies degradation. Uncertainty is reported as 95\% bootstrap confidence intervals (1,000 resamples).

\subsection{Task-Level Diagnostic Gap}

We formalize the task-dependent degradation captured by the framework. Let $\mathcal{T}_c$ denote a coarse diagnostic task (e.g., lesion type) and $\mathcal{T}_f$ a fine-grained task (e.g., pathology classification). The task-level diagnostic gap at a given fidelity operating point is:
\begin{equation}
\Delta_{\text{diag}} = \text{Pres}_{\mathcal{T}_c} - \text{Pres}_{\mathcal{T}_f},
\label{eq:diagnostic_gap}
\end{equation}
where $\text{Pres}_{\mathcal{T}}$ denotes any preservation metric (agreement, $\Delta$AUC, or $\overline{|\Delta p|}$) for task $\mathcal{T}$, evaluated on original and reconstructed inputs using the frozen probe. A positive $\Delta_{\text{diag}}$ indicates that reconstruction preserves coarse discriminative cues more reliably than fine-grained features. If $\Delta_{\text{diag}}$ varies with fidelity, the gap is controllable; if it remains positive even at high fidelity, fine-grained tasks are inherently more vulnerable to reconstruction.

\subsection{Explanation Preservation}

Decision preservation alone does not guarantee that reconstruction preserves the \emph{reasons} for a decision. We compute saliency maps for the probe's predicted class on the original and on each reconstruction using two gradient-based methods---HiResCAM~\cite{draelos2021hirescam} and Grad-CAM++~\cite{chattopadhay2018gradcampp}---at the final convolutional block. Similarity is measured by cosine similarity and SSIM of the attribution maps. Reporting two methods verifies that conclusions do not depend on a single attribution definition.

\section{Experimental Setup}
\label{sec:experiment}

\subsection{Dataset}

Experiments use lesion ROI crops from CBIS-DDSM~\cite{lee2017cbisddsm}, resized to $256\times256$ grayscale images. Patient-level splits are used throughout, yielding 2,494 training ROIs, 536 validation ROIs, and 520 held-out test ROIs (3,550 total). The dataset contains 1,880 calcification ROIs and 1,670 mass ROIs. For pathology classification, it includes 2,107 benign and 1,443 malignant ROIs overall.

The held-out test set contains 249 calcification ROIs (157 benign, 92 malignant) and 271 mass ROIs (148 benign, 123 malignant). Class imbalance is addressed using inverse-frequency weighting during probe training.

\subsection{Probe Training}

Each classifier is trained with AdamW (learning rate $10^{-4}$, weight decay $10^{-4}$), a cosine learning-rate schedule, and class-weighted cross-entropy for 10-25 epochs. Checkpoints are selected by validation balanced accuracy, followed by temperature scaling on the validation set. Held-out test performance establishes these models as valid evaluators: the lesion-type classifier reaches 0.915 accuracy; the mass-pathology classifier reaches 0.727 accuracy and 0.802 AUC; and the calcification-pathology classifier reaches 0.562 accuracy and 0.674 AUC. Although the calcification classifier performs better than random, it is the weaker one; therefore, mass-pathology results are treated as the primary evidence. After calibration, all evaluators are frozen for the reconstruction experiments.

\subsection{Calibration and Conformal Prediction}

Because the diagnostic gap framework uses the probe as a fixed measurement instrument, its calibration quality directly affects the reliability of preservation metrics---particularly the probability shift $\overline{|\Delta p|}$. We assess calibration using two post-hoc methods: temperature scaling~\cite{guo2017calibration}, which fits a single scalar on the validation logits, and Dirichlet calibration~\cite{kull2019dirichlet}, which fits a multiclass transform.

We further apply score-based conformal prediction~\cite{angelopoulos2021conformal} to construct prediction sets with finite-sample marginal coverage guarantees. For each target coverage level $(1-\alpha)$, conformal prediction produces a set of candidate labels whose size reflects the probe's uncertainty: a well-calibrated, confident probe yields small sets (near 1.0), while an uncertain probe yields larger sets (approaching the number of classes). This provides a model-free characterization of probe reliability that complements accuracy and AUC.

\subsection{Reconstructor Training}

All reconstructors are trained on train-split ROIs at $[-1,1]$ normalization with flip/rotation/affine augmentation at $256\times256$, conditioned on the lesion's class label. A separate reconstructor is trained per cascade stage.

The VQ-VAE-GAN and VAE-GAN use a two-phase schedule totaling 10 epochs: an initial phase minimizing $100 \cdot L_{\text{MSE}} + L_{\text{commit}} + 2 \cdot L_{\text{perc}}$, followed by PatchGAN adversarial refinement with the adversarial weight ramped from 0 to 0.5 and a halved generator learning rate. Checkpoints are selected on validation reconstruction MSE. The diffusion model is trained for 30 epochs with EMA weights used at inference. SDEdit strength is set to 0.5 for the primary evaluation.

All experiments are implemented in PyTorch with torchvision backbones. The diffusion model uses Hugging Face \texttt{diffusers}. Training uses a single instance of NVIDIA A100.

\section{Results}
\label{sec:results}

All numbers are computed on the held-out test split with the diagnostic probe frozen. The calcification probe is the weaker arm (test AUC 0.674 vs.\ 0.802 for mass), so mass and pooled results are treated as primary.

\subsection{Model Calibration}

\begin{figure}[t]
    \centering
    \includegraphics[width=\linewidth]{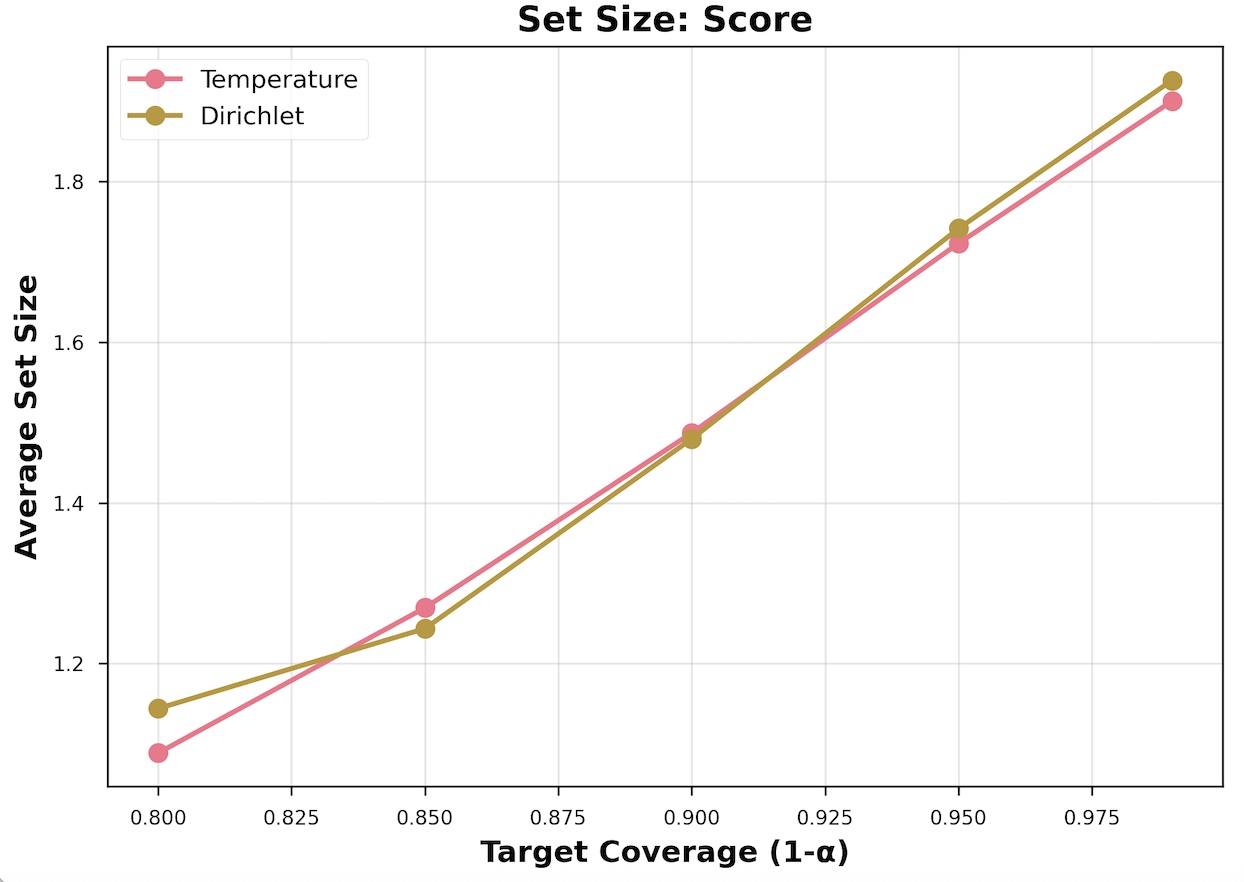}
    \caption{Average conformal prediction set size as a function of target coverage $(1-\alpha)$. The lesion-type classifier (Stage~1) produces the most efficient sets (size $\approx$1.0 at 90\% coverage), confirming strong calibration. Pathology model require larger sets, reflecting genuine diagnostic difficulty.}
    \label{fig:conformal_setsize}
\end{figure}

\begin{figure}[t]
    \centering
    \includegraphics[width=\linewidth]{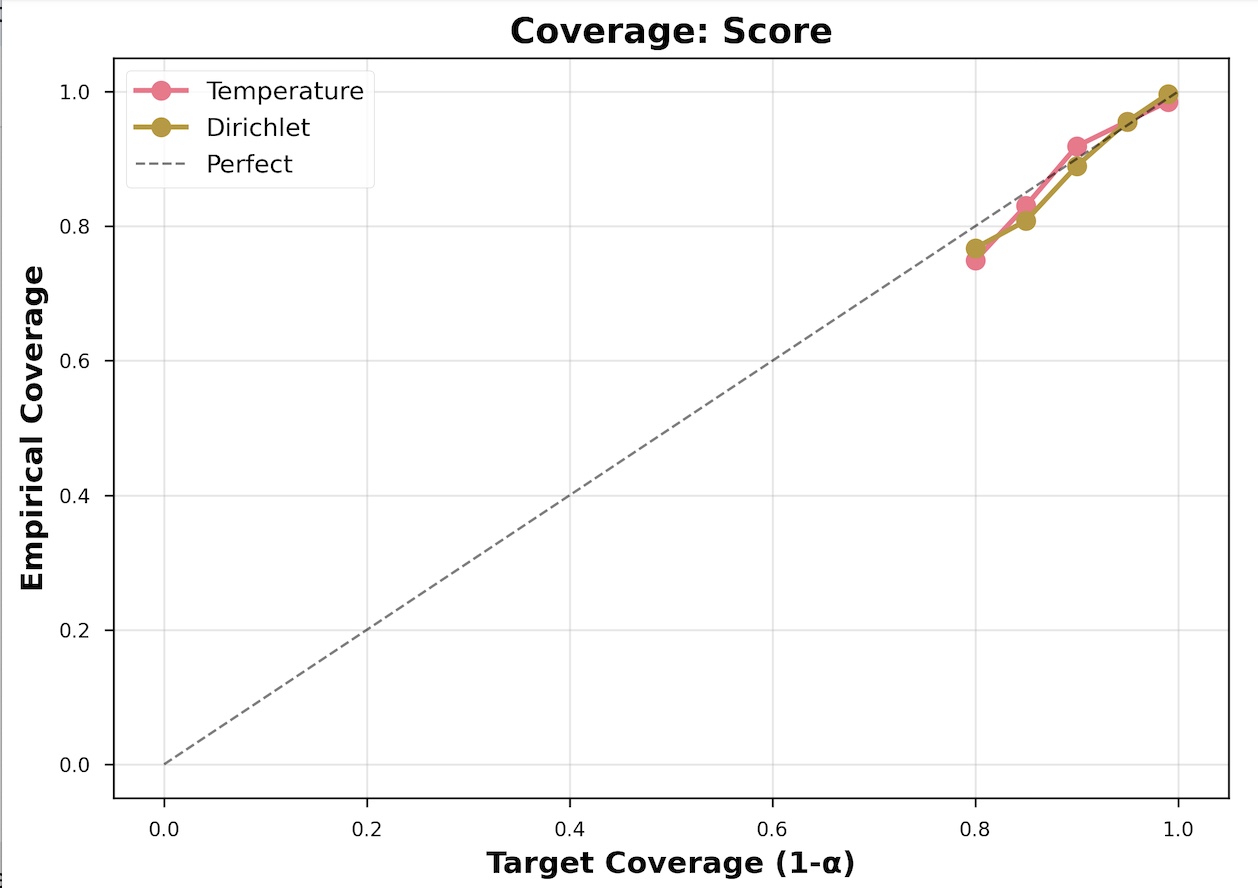}
    \caption{Empirical coverage as a function of target coverage $(1-\alpha)$ under score-based conformal prediction. Temperature scaling tracks the target more closely than Dirichlet calibration across all stages, providing tighter uncertainty estimates.}
    \label{fig:conformal_coverage}
\end{figure}

Conformal prediction set sizes (\Cref{fig:conformal_setsize}) confirm the probe hierarchy: the lesion-type produces near-singleton sets (average size 0.83--1.25 across 80--97.5\% target coverage), indicating high confidence with valid coverage. The pathology probes require progressively larger sets, consistent with the lower AUC on these harder tasks. Temperature scaling tracks target coverage more closely than Dirichlet calibration without premature saturation (\Cref{fig:conformal_coverage}), providing more efficient uncertainty estimates. This calibration analysis validates them as a reliable measurement instrument: its confidence is well-aligned with its actual performance, ensuring that the probability shift metric $\overline{|\Delta p|}$ captures genuine reconstruction-induced degradation rather than calibration artifacts.

\subsection{The Fidelity Ladder Spans a Wide Range}

The three reconstructors occupy clearly separated points on the fidelity axis (\Cref{tab:decision}). The autoencoders are near-lossless: VAE-GAN achieves the highest fidelity (LPIPS 0.029, SSIM 0.989), with VQ-VAE-GAN close behind (LPIPS 0.062, SSIM 0.980). Diffusion under SDEdit sits far lower (LPIPS 0.584, SSIM 0.408)---roughly a twenty-fold larger perceptual distance. This range enables reading diagnostic preservation against fidelity rather than against model identity.

\subsection{Decision Preservation Tracks Fidelity}

\begin{table*}[t]
\centering
\caption{Decision preservation across the fidelity ladder with 95\% bootstrap confidence intervals (1,000 resamples). $\Delta$AUC and $\Delta$Acc report changes from the original probe performance. At autoencoder fidelity, diagnostics are preserved ($\Delta$AUC CIs span zero); at diffusion fidelity, they degrade sharply (CIs exclude zero).}
\label{tab:decision}
\begin{tabular}{llcccccl}
\toprule
Stage & Reconstructor & LPIPS$\downarrow$ & SSIM$\uparrow$ & Agreement & $\Delta$Acc & $\Delta$AUC [95\% CI] & $\overline{|\Delta p|}$ [95\% CI] \\
\midrule
\multirow{3}{*}{\shortstack[l]{Lesion Type\\(n=520)}}
 & VQ-VAE-GAN & 0.062 & 0.980 & 0.965 & 0.000 & $-$0.007 [$-$0.012, $-$0.002] & 0.046 [0.038, 0.054] \\
 & VAE-GAN    & 0.029 & 0.989 & 0.975 & 0.000 & $-$0.002 [$-$0.005, +0.002] & 0.030 [0.024, 0.035] \\
 & Diffusion  & 0.584 & 0.408 & 0.692 & $-$0.188 & $-$0.105 [$-$0.136, $-$0.074] & 0.304 [0.273, 0.333] \\
\midrule
\multirow{3}{*}{\shortstack[l]{Pathology\\Pooled (n=520)}}
 & VQ-VAE-GAN & 0.062 & 0.980 & 0.865 & 0.000 & $-$0.001 [$-$0.025, +0.022] & 0.082 [0.075, 0.088] \\
 & VAE-GAN    & 0.029 & 0.989 & 0.888 & +0.012 & +0.004 [$-$0.017, +0.026] & 0.076 [0.069, 0.083] \\
 & Diffusion  & 0.584 & 0.408 & 0.604 & $-$0.188 & $-$0.253 [$-$0.307, $-$0.202] & 0.218 [0.205, 0.231] \\
\midrule
\multirow{3}{*}{\shortstack[l]{Mass\\Pathology (n=271)}}
 & VQ-VAE-GAN & 0.047 & 0.983 & 0.838 & 0.000 & +0.001 [$-$0.032, +0.033] & 0.103 [0.093, 0.114] \\
 & VAE-GAN    & 0.030 & 0.988 & 0.875 & +0.007 & $-$0.005 [$-$0.035, +0.023] & 0.098 [0.087, 0.109] \\
 & Diffusion  & 0.634 & 0.213 & 0.535 & $-$0.273 & $-$0.341 [$-$0.422, $-$0.261] & 0.239 [0.222, 0.256] \\
\midrule
\multirow{3}{*}{\shortstack[l]{Calc\\Pathology (n=249)}}
 & VQ-VAE-GAN & 0.078 & 0.977 & 0.896 & 0.000 & $-$0.001 [$-$0.033, +0.033] & 0.058 [0.052, 0.065] \\
 & VAE-GAN    & 0.028 & 0.989 & 0.904 & +0.016 & $-$0.004 [$-$0.032, +0.024] & 0.052 [0.046, 0.057] \\
 & Diffusion  & 0.530 & 0.621 & 0.679 & $-$0.096 & $-$0.158 [$-$0.243, $-$0.074] & 0.194 [0.178, 0.213] \\
\bottomrule
\end{tabular}
\end{table*}

At autoencoder fidelity, the diagnosis is preserved to within measurement noise (\Cref{tab:decision}). For both VQ-VAE-GAN and VAE-GAN, pathology $\Delta$AUC is within $\pm$0.005 of the original probe and accuracy is effectively unchanged; bootstrap 95\% CIs span zero in all cases, confirming the effect is not statistically significant. Diffusion is a qualitatively different regime: pooled pathology AUC falls by 0.253 (CI: [$-$0.307, $-$0.202]), and on mass it drops by 0.341 to 0.461---\emph{below chance}---indicating reconstruction so corrupts the lesion that the probe is systematically misled rather than merely noisier. All diffusion CIs exclude zero, confirming some statistically significant degradation.

\textbf{Lesion-type preservation.} Lesion-type discrimination is more resilient to reconstruction degradation than fine-grained pathology classification across all fidelity levels. Autoencoder reconstructions retain 96.5--97.5\% agreement for lesion type, compared with 84--90\% for pathology, while diffusion retains 69.2\% agreement. Nevertheless, the lesion-type AUC still declines under diffusion ($\Delta$AUC $=-0.105$), indicating that coarse diagnostic information is not fully preserved at low fidelity. Rather, it degrades less severely than fine-grained pathology information. This difference in degradation across diagnostic tasks is what the diagnostic gap metric is designed to quantify.

The continuous probability shift mirrors the fidelity ordering: $\overline{|\Delta p|}$ is two to three times larger under diffusion (0.218 pooled pathology) than under the autoencoders (0.076--0.082). The signed shift is positive, indicating SDEdit biases predictions toward malignant rather than scattering symmetrically.

Prediction agreement is the least informative measure: even high-fidelity rungs flip 10--16\% of pathology predictions (agreement 0.84--0.90), yet their AUC is intact---the flips fall on borderline cases and do not change the probe's ability to rank malignant above benign. $\Delta$AUC and probability shift, not raw agreement, separate ``preserved'' from ``degraded.''

\begin{figure}[t]
    \centering
    \includegraphics[width=\columnwidth]{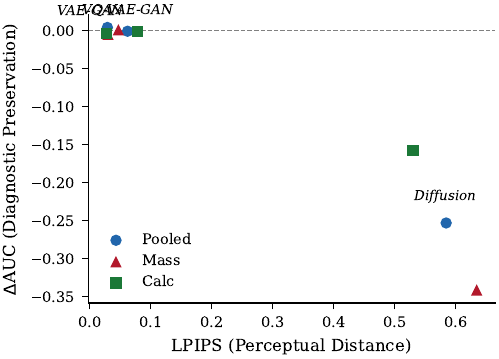}
    \caption{Decision preservation ($\Delta$AUC) as a function of reconstruction fidelity (LPIPS). At autoencoder fidelity (low LPIPS), AUC is preserved within $\pm$0.005. At diffusion fidelity (high LPIPS), AUC degrades sharply, with mass pathology falling below chance.}
    \label{fig:fidelity_preservation}
\end{figure}

\Cref{fig:fidelity_preservation} shows the relationship between LPIPS and $\Delta$AUC across all subsets and reconstructors. The diagnostic gap is clearly fidelity-dependent: autoencoder operating points cluster near $\Delta$AUC~$= 0$, while diffusion operating points fall sharply below.

\subsection{Explanation Preservation Tracks Fidelity}

\begin{table}[t]
\centering
\caption{Attribution preservation across the fidelity ladder. Cosine similarity and SSIM between saliency maps on original and reconstructed ROIs. Both HiResCAM and Grad-CAM++ show the same ordering.}
\label{tab:attribution}
\begin{tabular}{lcccc}
\toprule
 & \multicolumn{2}{c}{HiResCAM} & \multicolumn{2}{c}{Grad-CAM++} \\
\cmidrule(lr){2-3} \cmidrule(lr){4-5}
Reconstructor & Cos & SSIM & Cos & SSIM \\
\midrule
VAE-GAN    & 0.835 & 0.635 & 0.899 & 0.762 \\
VQ-VAE-GAN & 0.782 & 0.585 & 0.869 & 0.718 \\
Diffusion  & 0.511 & 0.367 & 0.713 & 0.526 \\
\bottomrule
\end{tabular}
\end{table}

\begin{figure}[t]
    \centering
    \includegraphics[width=\columnwidth]{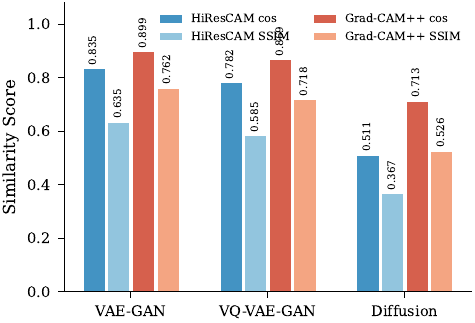}
    \caption{Attribution preservation across the fidelity ladder. Both HiResCAM and Grad-CAM++ show the same ordering: autoencoders preserve attention substantially better than diffusion, and the effect is consistent across similarity metrics.}
    \label{fig:attribution}
\end{figure}

The same fidelity ordering holds for \emph{where} the probe looks (\Cref{tab:attribution}, \Cref{fig:attribution}). Saliency-map similarity is high for the autoencoders and substantially lower for diffusion. The ranking is identical under HiResCAM and Grad-CAM++, confirming the effect is not an artifact of one attribution definition. Grad-CAM++ reads higher in absolute terms because its maps are smoother, but the gap and order are the same under both methods.

A key finding distinguishes explanation fidelity from decision fidelity. At autoencoder fidelity, the \emph{decision} is preserved almost perfectly ($\Delta$AUC~$\approx 0$) while the \emph{attribution} is only moderately preserved (SSIM 0.59--0.76)---reconstruction shifts where the probe looks even when it does not change what the probe decides. Explanation preservation is therefore the more sensitive diagnostic indicator: it discriminates VAE-GAN from VQ-VAE-GAN (VAE-GAN preserves attention slightly better, consistent with its lower LPIPS) where decision metrics saturate. At diffusion fidelity, both collapse together.

This decision--explanation dissociation has practical implications for clinical AI. A pipeline that preserves the correct diagnosis but shifts the classifier's attention to different tissue regions may yield correct labels while undermining interpretability---a concern in regulatory settings where explanations are expected to accompany predictions. The diagnostic gap framework surfaces this distinction automatically by measuring both axes.

\subsection{Fidelity Governs the Diagnostic Gap}

The two preservation curves together answer a central methodological question: is the diagnostic gap an intrinsic artifact of reconstruction, or a function of fidelity? The evidence supports the latter. At VQ-VAE-GAN fidelity, both the decision ($\Delta$AUC within $\pm$0.005) and the explanation are substantially preserved, so the gap at that operating point is small. The gap grows only as fidelity degrades, and it grows on both axes in step with measured LPIPS and SSIM. A high-fidelity reconstructor incurs little diagnostic cost; a low-fidelity one can push classification below chance.

\subsection{SDEdit Strength Sweep}

\begin{figure}[t]
    \centering
    \includegraphics[width=\columnwidth]{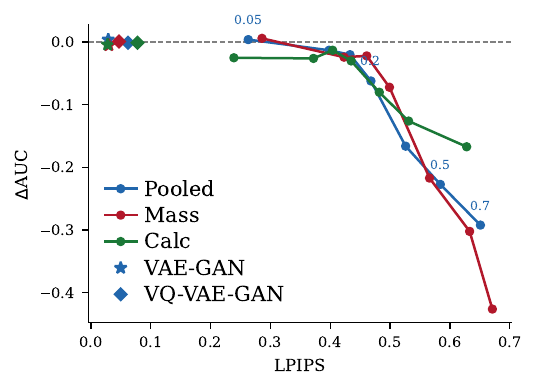}
    \caption{Dose-response curve: $\Delta$AUC as a function of LPIPS across SDEdit strengths (0.05--0.7) and autoencoder baselines. Degradation increases monotonically with perceptual distance across all subsets. Mass pathology is most vulnerable, dropping below chance at strength $\geq$0.5.}
    \label{fig:sweep}
\end{figure}

\begin{table*}[t]
\centering
\caption{SDEdit strength sweep: decision preservation across seven strength levels for pooled pathology (n=520). Autoencoder baselines shown for reference. $\Delta$AUC degrades monotonically with increasing strength; even at minimal strength (0.05), LPIPS exceeds all autoencoder operating points.}
\label{tab:sweep}
\begin{tabular}{llccccccc}
\toprule
Set & Point & Strength & LPIPS$\downarrow$ & SSIM$\uparrow$ & Agree & $\Delta$Acc & $\Delta$AUC & $\overline{|\Delta p|}$ \\
\midrule
\multirow{9}{*}{Pooled}
 & VAE-GAN    & ---  & 0.029 & 0.989 & 0.888 & +0.012 & +0.004 & 0.076 \\
 & VQ-VAE-GAN & ---  & 0.062 & 0.980 & 0.865 & 0.000  & $-$0.001 & 0.082 \\
 \cmidrule(lr){2-9}
 & SDEdit     & 0.05 & 0.263 & 0.866 & 0.735 & $-$0.042 & +0.004  & 0.181 \\
 & SDEdit     & 0.10 & 0.398 & 0.700 & 0.660 & $-$0.090 & $-$0.013 & 0.207 \\
 & SDEdit     & 0.15 & 0.433 & 0.646 & 0.663 & $-$0.071 & $-$0.020 & 0.214 \\
 & SDEdit     & 0.20 & 0.468 & 0.591 & 0.660 & $-$0.094 & $-$0.062 & 0.209 \\
 & SDEdit     & 0.35 & 0.526 & 0.501 & 0.613 & $-$0.117 & $-$0.166 & 0.220 \\
 & SDEdit     & 0.50 & 0.584 & 0.408 & 0.604 & $-$0.169 & $-$0.227 & 0.214 \\
 & SDEdit     & 0.70 & 0.651 & 0.342 & 0.535 & $-$0.127 & $-$0.292 & 0.225 \\
\bottomrule
\end{tabular}
\end{table*}

To test whether preservation tracks reconstruction fidelity consistently, we sweep SDEdit strength across seven levels (0.05, 0.1, 0.15, 0.2, 0.35, 0.5, 0.7), tracing preservation from near-identity reconstruction to aggressive regeneration (\Cref{fig:sweep}). The results confirm monotonic degradation: as strength increases (and LPIPS rises from 0.26 to 0.65 pooled), $\Delta$AUC decreases steadily from near zero to $-$0.292, with no recovery or plateau. Mass pathology is consistently the most vulnerable subset, crossing below chance ($\Delta$AUC $< -0.3$) at strength 0.5 and reaching $-$0.426 at strength 0.7. Calcification pathology degrades more gradually ($\Delta$AUC from $-$0.025 to $-$0.167), consistent with calcifications being morphologically more distinctive.

A key finding is the fidelity gap between reconstruction paradigms. Even at the gentlest tested strength (0.05), the diffusion model produces LPIPS $\approx$0.26---already four times higher than VQ-VAE-GAN (0.062) and nine times higher than VAE-GAN (0.029). At this strength, $\Delta$AUC is near zero (pooled: +0.004), indicating that the diagnostic signal is preserved despite the perceptual distance. Degradation becomes significant only at strength $\geq$0.2 (pooled $\Delta$AUC $= -$0.062). This establishes that SDEdit reconstruction incurs a baseline fidelity cost even at minimal strength, and that diagnostic preservation degrades smoothly as that cost increases.

\section{Discussion}
\label{sec:discussion}

\subsection{Framework Utility}

The diagnostic gap framework adds three capabilities beyond standard reconstruction evaluation.

\textbf{Isolation of reconstruction effects.} Freezing the evaluators ensures that changes in output are attributable to reconstruction rather than model adaptation. A retrained classifier could learn to compensate for reconstruction artifacts and hide genuine information loss. Using the same frozen evaluator at every fidelity level therefore enables a controlled comparison across reconstructors.

\textbf{Complementary decision and explanation measures.} Decision and explanation fidelity capture different forms of degradation. Attribution similarity distinguishes VAE-GAN from VQ-VAE-GAN when decision metrics are nearly identical, whereas decision metrics clearly separate the autoencoder and diffusion regimes. Measuring both provides a more complete view of pipeline behavior than either measure alone.

\textbf{A fidelity-based comparison axis.} Relating preservation to measured fidelity (LPIPS and SSIM), rather than to model identity alone, produces a reusable dose--response curve. New reconstructors can be placed on the same fidelity axis and compared without treating each architecture as a separate evaluation setting.

\textbf{Limits of prediction agreement.} Prediction agreement remains relatively high (0.84--0.90) even when AUC is unchanged, but it does not distinguish harmless borderline prediction flips from systematic changes in diagnostic ranking. The bootstrap intervals in \Cref{tab:decision} make this distinction explicit: Bootstrap confidence intervals show that autoencoder-related changes in AUC are indistinguishable from zero, whereas diffusion produces a statistically reliable AUC decline. Prediction agreement alone cannot distinguish between these cases because it only counts whether the final class label changes; it does not capture shifts in confidence or ranking that may reduce AUC even when many predicted labels remain unchanged.

\subsection{Implications for Reconstruction Choice}

The fidelity ladder shows that diagnostic preservation depends primarily on achieved reconstruction fidelity, not simply on the reconstruction architecture. Both VAE-GAN and VQ-VAE-GAN preserve diagnostic performance at high fidelity despite their different latent representations. In contrast, diffusion reconstruction at SDEdit strength 0.5 operates at substantially lower fidelity and degrades diagnostic performance. The more useful question is therefore not `which architecture preserves diagnostics?'' but `at what fidelity does diagnostic preservation begin to fail?''

The mass-pathology results illustrate this effect clearly. Diffusion reduces AUC to 0.461, with $\Delta$AUC $=-0.342$ (95\% CI: [$-0.422$, $-0.261$]), below the 0.5 chance level. This suggests that the reconstruction does more than add random noise: it changes the ranking of benign and malignant cases in a systematic way. One plausible explanation is the loss of fine textural features that help distinguish benign from malignant masses. Calcification pathology shows a smaller but still significant decline ($\Delta$AUC $=-0.158$, CI excludes zero), consistent with calcifications retaining more visually distinctive structure under reconstruction.

\textbf{Discrete versus continuous latent spaces.} VQ-VAE-GAN and VAE-GAN achieve nearly identical decision preservation ($\Delta$AUC within $\pm 0.005$). However, VAE-GAN attains slightly better image fidelity (LPIPS 0.029 versus 0.062) and better attribution preservation (HiResCAM cosine 0.835 versus 0.782). This suggests that vector quantization introduces minor spatial changes that are visible in explanation metrics but are not large enough to alter diagnostic decisions at this fidelity level. For ROI-level reconstruction, continuous-latent models may therefore offer a modest explanation-fidelity advantage without sacrificing decision fidelity.

\subsection{Task-Dependent Degradation}

Coarse diagnostic tasks are consistently more robust to reconstruction degradation than fine-grained ones. Lesion-type classification (mass versus calcification) retains 96.5--97.5

The diagnostic gap metric $\Delta_{\text{diag}}$ (\Cref{eq:diagnostic_gap}) quantifies this difference. Using prediction agreement as the preservation measure, the gap remains positive across the fidelity ladder: 0.100 for VQ-VAE-GAN, 0.087 for VAE-GAN, and 0.088 for diffusion at strength 0.5. The stable 9--10 percentage-point gap suggests that the difference is not limited to one reconstruction method. High coarse-task performance can therefore overstate the diagnostic reliability of a pipeline when fine-grained pathology information is less well preserved.

\subsection{Scope and Limitations}

\textbf{ROI-only evaluation.} This study isolates reconstruction from localization. The complete weakly supervised pipeline also includes attribution-based lesion localization on full-field mammograms, which can introduce additional information loss. The framework can be extended by treating localization as an upstream stage, allowing future work to measure how localization error interacts with reconstruction fidelity. While Grad-CAM-based localization can recover lesion-relevant spatial support, the combined effect of localization and reconstruction errors remains untested.

\textbf{Diffusion model capacity.} The diffusion model is a moderately sized U-Net trained from scratch on CBIS-DDSM. Larger or pretrained diffusion models may achieve higher fidelity at the same SDEdit strength and would likely shift diffusion operating points toward the high-fidelity end of the curve. However, even the lowest tested strength (0.05) produces LPIPS of approximately 0.26, substantially higher than the autoencoder range ($\leq 0.08$). The observed fidelity--preservation trend suggests that the main conclusion should remain the same: diagnostic degradation tracks achieved fidelity. Testing stronger diffusion models would verify this directly.

\textbf{Dataset and evaluator scope.} The dataset contains less than four-thousand images and a relatively small test set. The calcification-pathology evaluator is weaker than the mass-pathology evaluator (AUC 0.674 versus 0.802), which limits the precision of conclusions for calcifications. Evaluation on additional datasets, such as such as INbreast~\cite{moreira2012inbreast} and VinDr-Mammo~\cite{nguyen2023vindrmammo}, would test whether the observed fidelity--preservation relationship generalizes across populations, scanners, and acquisition settings.

\textbf{Attribution methods.} We evaluate two gradient-based attribution methods, HiResCAM and Grad-CAM++. Perturbation-based methods, such as RISE~\cite{petsiuk2018rise} and occlusion sensitivity~\cite{zeiler2014visualizing}, and concept-based methods, like TCAV~\cite{kim2018tcav}, may show different sensitivity patterns. The framework can accommodate these methods as long as they produce a spatial explanation map that can be compared across original and reconstructed inputs.

\subsection{Future Directions}

Several extensions follow naturally from this work. First, adding localization as an upstream stage would enable end-to-end evaluation of weakly supervised pipelines and quantify how localization error compounds with reconstruction loss. Second, applying the framework to other modalities, including chest X-ray, histopathology, CT, and MRI, would test the generality of the task-dependent degradation pattern. Third, explanation preservation could be incorporated into reconstructor training, encouraging models to preserve diagnostic attention rather than optimizing only pixel-level or perceptual similarity. Finally, combining the framework with selective prediction or conformal methods could identify whether reconstruction-induced degradation is concentrated in cases where the classifier is already uncertain.

\section{Conclusion}
\label{sec:conclusion}

We presented the diagnostic gap framework, a practical evaluation tool for assessing how reconstruction fidelity affects diagnostic information in multi-stage AI pipelines. Using a fidelity ladder of three reconstructors on mammography ROIs, we showed that the diagnostic gap is fidelity-dependent, not intrinsic: at autoencoder fidelity (LPIPS $\leq$ 0.078), both decision and explanation are preserved; at diffusion fidelity (LPIPS 0.584), both collapse. Explanation preservation is the more sensitive indicator, discriminating reconstructors where decision metrics saturate. The framework is architecture-agnostic and provides reusable components for researchers building weakly supervised medical AI systems.

\bibliographystyle{ieeetr}
\bibliography{references}

\end{document}